\documentclass[journal]{IEEEtran}
\ifCLASSINFOpdf
\else
\fi
\usepackage{url}

\usepackage{graphicx}

\begin{document}
%
\title{Intensity Video Guided 4D Fusion for\\Improved Highly Dynamic 3D Reconstruction}
%
%
%

\author{Jie~Zhang, 
       Christos~Maniatis, 
       Luis~Horna，
       and~Robert~B.~Fisher 
\thanks{Jie Zhang is with the School of Instrumentation Science and Opto-electronics Engineering, Beihang University, Beijing 100191, China}
\thanks{Jie Zhang, Luis~Horna and Robert B. Fisher are with the School of Informatics, the University of Edinburgh, EH8 9AB, UK}
\thanks{Christos~Maniatis was with the School of Mathematics, the University of Edinburgh, EH8 9AB, UK}
\thanks{*Corresponding author: zhangjie09@buaa.edu.cn}}

\maketitle

\begin{abstract}
The availability of high-speed 3D video sensors has greatly facilitated 3D shape acquisition of dynamic and deformable objects, but high frame rate 3D reconstruction is always degraded by spatial noise and temporal fluctuations. This paper presents a simple yet powerful intensity video guided multi-frame 4D fusion pipeline. Temporal tracking of intensity image points (of moving and deforming objects) allows registration of the corresponding 3D data points, whose 3D noise and fluctuations are then reduced by spatio-temporal multi-frame 4D fusion. We conducted simulated noise tests and real experiments on four 3D objects using a 1000 fps 3D video sensor. The results demonstrate that the proposed algorithm is effective at reducing 3D noise and is robust against intensity noise. It outperforms existing algorithms with good scalability on both stationary and dynamic objects.
\end{abstract}

\begin{IEEEkeywords}
high-speed 3D video sensor, multi-frame 4D fusion, intensity tracking, dynamic object, noise reduction
\end{IEEEkeywords}

%
\IEEEpeerreviewmaketitle

\section{Introduction}
\IEEEPARstart{T}{hree} dimensional shape acquisition of highly dynamic and deformable objects is an increasingly active research topic in computer vision with the development of high-speed 3D video sensors \cite{xiao2011performance,tabata2015high}. It is a fundamental and critical prerequisite of numerous applications, such as dynamic face recognition \cite{zhang2014bp4d}, action and behavior perception \cite{SPaction2013,SPmotion2015}, object deformation analysis, etc. However, the 3D sequences from high-speed 3D video sensors usually suffer from serious spatial noise and temporal fluctuations that degrade the performance of 3D reconstruction. The inaccuracy of the high frame rate 3D sequence is caused by multiple general factors, including calibration error, non-uniform illumination, the surface property and motion of scenes or objects, etc. Additionally, resulting from the sensor technology, there are a small number of out-of-sync pixels that produce spatial noise and temporal fluctuations in the 3D sequence. Therefore, denoising high frame rate 3D/depth sequences and thus improving the performance of 3D dynamic and deformable shape acquisition are of significant value.

3D/depth noise characterization and models \cite{mallick2014characterizations,khoshelham2012accuracy,yu2012shadow,nguyen2012modeling,park2012spatial} provide an important basis for boosting 3D reconstruction performance. Noise in a 3D/depth image can be generally characterized into three types (spatial, temporal and interference noise) with corresponding theoretical and empirical noise models. Existing 3D/depth image improvement methods mainly focus on reducing spatial axial and lateral noise, smoothing temporal fluctuations and filling non-measured pixels. They are performed either using a single image (adaptive Gaussian filter (Ad-GF) \cite{nguyen2012modeling}, adaptive bilateral filter (Ad-BF) \cite{chen2012depth}) or multiple registered images (KinectFusion \cite{izadi2011kinectfusion}, imaging burst \cite{hasinoff2016burst}). The multi-view 3D registration based methods \cite{izadi2011kinectfusion,SPregistration} are helpful in smoothing the 3D data and thus improving the 3D reconstruction quality, while the performance of the methods on dynamic or deformable objects is still limited. To address this, there are existing algorithms using motion/temporal information in point-based fusion \cite{keller2013real} or filtering, such as velocity-based adaptive threshold filter (Ad-TF) \cite{essmaeel2012temporal}, spatial-temporal divisive normalized bilateral filter (DNBF) \cite{fu2012kinect}, constrained temporal averaging (TA) \cite{wasza2011real}). Those algorithms improve the 3D reconstruction of dynamic scenes while are only based on depth information. On the other hand, depth-intensity based 3D/depth noise reduction methods (adaptive joint bilateral filter (Ad-JBF) \cite{camplani2013depth}, guided filter \cite{he2013guided}, non-causal spatio-temporal median filter (ST-MF) \cite{matyunin2011temporal}) and multi-sensor systems \cite{yang2015evaluating} have been used for boosting the quality of 3D reconstruction. However, due to the limited reconstruction quality of high-speed 3D video sensors, denoising high frame rate sequences is still an open issue.

In this paper, we focus on intensity tracking guided 4D fusion for boosting the 3D reconstruction from high-speed 3D video sensors. The core idea behind the method is that the intensity data of consecutive images can be aligned by a temporal "stereo" matching algorithm, and then the corresponding 3D point data can be fused in the spatio-temporal domain to reduce the 3D data noise and fluctuations. Our contributions are: 

(1) a generic intensity tracking guided multi-frame 4D fusion model that integrates spatial intra-frame filtering and temporal inter-frame fusion. (Sec. \ref{secpipeline})

(2) a simple yet powerful pipeline for boosting the 3D reconstruction of dynamic and deformable objects. (Sec. \ref{secfusion})

(3) we demonstrate these by denoising 3D sequences of stationary, dynamic and deformable objects from a $1000$ fps 3D video sensor. (Sec. \ref{secresults})

\begin{figure*}
\centering
\includegraphics[width= 14cm]{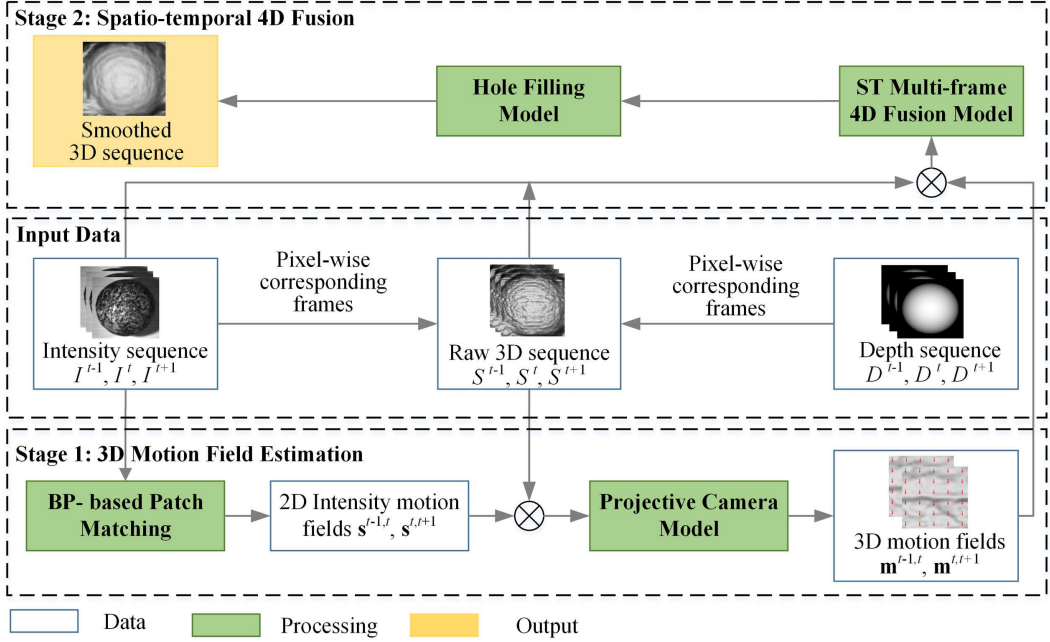}
\caption{The system framework (using 3 consecutive frames as an example).}
\label{figoverview}
\end{figure*}

\section{Proposed Pipeline}
\label{secpipeline}
The proposed system framework (Fig. \ref{figoverview}) has 2 main stages: (1) intensity tracking guided 3D motion field estimation; (2) spatio-temporal multi-frame 4D fusion. 
Given a 3D sequence $S^t=\{{\bf{p}}^t_i \in {\mathcal{R}}^3\}$ with pixel-wise registered intensity images $I^t=\{a^t_i \in \mathcal{R}\}$ and depth images $D^t=\{d^t_i \in \mathcal{R}\}$, in the first stage, dense tracking is performed on the intensity sequence $I^t$ using a belief propagation based patch matching algorithm \cite{besse2014pmbp} for optical flow, which obtains continuous intensity motion fields. Then, using the projective camera model, the pixel-wise 3D motion fields of the registered 3D sequence $P^t$ can be estimated by leveraging the intensity motion fields. In the second stage, piecewise spatio-temporal multi-frame 4D fusion is performed on the 3D sequence using the 3D motion fields. Since the rejected outliers in the 3D motion fields result in holes in the fused 3D sequence, we perform gradient-directed hole filling to repair them. Finally, an improved 3D sequence can be obtained. More details are given in Section \ref{secfusion}.

\section{Intensity Tracking Guided 4D Fusion}
\label{secfusion}
This section details the intensity tracking guided 3D motion field estimation and the spatio-temporal multi-frame 4D fusion model.

\subsection{Intensity-guided 3D Motion Field Estimation}
For a dynamic 3D object, assume that each intensity image point in $n$ consecutive frames is trackable in the temporal domain. Dense tracking is performed on the pixel-wise registered intensity sequence $I^t$ using a particle belief propagation method. This will give a motion field $\{\mathbf{s}^{t,t+1} \in {\mathcal{R}}^2\}$ between each pair of consecutive 2D intensity frames $I^t, I^{t+1}$. Then, the pixel-wise continuous intensity motion fields give pixel-wise correspondences in the registered depth frames $D^t$. We iterate the correspondence so each point has a known position $\mathbf{p}_i^t$ in the 3D frame $S^t$. 

The intensity correspondence field $\mathbf{s}^{t,t+1}$ is obtained by minimizing an objective function that combines a unary term evaluating point similarity and a pairwise term for piecewise smoothness as:
\begin{equation}
\scalebox{0.86}{$ \{{\bf{\hat s}}_i^{t,t + 1}\}  = \mathop {\arg \min }\limits_{ \{{\bf{s}}_i^{t,t + 1}\}} \sum\limits_i {({\psi}_1 ({\bf{s}}_i^{t,t + 1} ) + \sum\limits_{n \in N_I(i)} {{\psi}_2 ({\bf{s}}_i^{t,t + 1} ,{\bf{s}}_n^{t,t + 1} )} )} $}
\end{equation}

$N_I(i)$ are the neighbors of 2D intensity pixel $a_i^t$ in frame $I^t$. ${\psi}_1 ({\bf{s}}_i^{t,t+1})$ is the unary term that represents the discrepancy of a pair of corresponding 2D intensity patches centered on the corresponding points ${a}_i^t, {a}_i^{t+1}$ in the consecutive frames $I^t, I^{t+1}$. ${{\psi}_2 ({\bf{s}}_i^{t,t + 1}, {\bf{s}}_n^{t,t + 1} )}$ is a smoothness term to regularize the correspondence field, which can be optimized by minimizing the message (smoothness error) passed by the neighboring intensity patch $n$ to the patch $i$. 

According to the projective camera model, the point ${\bf{p}}_i^t$ in the 3D frame $S^t$ can be expressed as
\begin{equation}
{\bf{p}}_i^t  = d_i^t \left[ {\begin{array}{*{20}c}
   {f_x ^{ - 1} (x_i^t  - u_0 )} & {f_y ^{ - 1} (y_i^t  - v_0 )} & 1  \\
\end{array}} \right]
\end{equation}
where $f_x, f_y, u_0, v_0$ are the calibration parameters (focal length and centers) of the camera,  $d_i^t $ is the depth value, and $ x_i^t ,y_i^t $ are intensity pixel coordinates.

With ${\bf{s}}_i^{t,T} = [s_{ix}^{t,T} ,s_{iy}^{t,T}]$, $x_i^{T}  = x_i^t  + s_{ix}^{t,T} $ and $y_i^{T} = y_i^t  + s_{iy}^{t,T}$, the 3D correspondence vector ${\bf{m}}_i^{t,T}$ for the point $i$ from frame $S^t$ to frame $S^T$ can be estimated by:
\begin{equation}
\scalebox{0.8}{${\bf{m}}_i^{t,T}  = \left[ {\begin{array}{*{20}c}
   {f_x ^{ - 1} (x_i^t  - u_0 )(d_i^{T}  - d_i^t ) + f_x ^{ - 1} d_i^{T} s_{ix}^{t,T} }  \\
   {f_y ^{ - 1} (y_i^t  - v_0 )(d_i^{T}  - d_i^t ) + f_y ^{ - 1} d_i^{T} s_{iy}^{t,T} }  \\
   {(d_i^{T}  - d_i^t )}  \\
\end{array}} \right]$}
\end{equation}

By tracking from frame to frame, we can link the intensity image point $a_i^t$ to its 3D position ${\bf{p}}_i^t$ in all the frames.

\subsection{Spatio-temporal Multi-frame 4D Fusion}
Given $n$ consecutive 3D frames, we seek to fuse them into one frame using the continuous 3D motion fields for piecewise spatio-temporal smoothness. Firstly, the outliers in each 3D motion field are removed by verifying pairwise forward and backward motion vectors with a threshold constraint. Specifically, for a pair of 3D motion vectors ${\bf{m}}_i^{t,t+1}$ (or expressed as ${\bf{m}}^{t,t+1}(x_i,y_i,z_i)$) and ${\bf{m}}_{i}^{t + 1,t}$ between a pair of corresponding points $\{{\bf{p}}_i^t,{\bf{p}}_i^{t+1}\}$ in the frame $S^t$ and $I^{t+1}$, the sum of the vectors should be smaller than a threshold $\vartheta$ (in practice, we choose $\vartheta = 2$ pixels) as: 
\begin{equation}
\scalebox{0.74}{$
\left\| {{\bf{m}}^{t,t + 1}(x_i,y_i,z_i)  + {\bf{m}}^{t + 1,t} (x_i + {\bf{m}}_{ix}^{t,t + 1} ,y_i + {\bf{m}}_{iy}^{t,t + 1} ,z_i + {\bf{m}}_{iz}^{t,t + 1} )} \right\| < \vartheta $}
\end{equation}

The 3D motion vectors that satisfy the threshold constraint are accepted as reasonable motion vectors.

The piecewise spatio-temporal 4D fusion performed on consecutive 3D frames can be expressed as
\begin{equation}
\scalebox{0.71}{${\bf{\hat p}}_i^t  = \frac{1}{{\kappa _i }}\sum\limits_{T \in N_t(t)}^{} {\nu _i^{t,T} f(t,T)\left\{ {\left[ {\frac{1}{{\kappa _i^T }}\sum\limits_{n \in N(i)} {d({\bf{p}}_i^T ,{\bf{p}}_n^T )g(I_i^T ,I_n^T ){\bf{p}}_n^T } } \right] - {\bf{m}}_i^{t,T} } \right\}}$}
\end{equation}

In the internal summation, ${\bf{p}}_n^{T}(n \in N(i))$ is a set of neighbors of the point $i$ in the frame $S^T$. $d({\bf{p}}_i^T ,{\bf{p}}_n^T ) = e^{-\left\| {{\bf{p}}_i^T  - {\bf{p}}_n^T } \right\| ^2/2\delta _{d}^2 }$ and $g(I_i^T ,I_n^T ) = e^{-\left| {I_i^T  - I_n^T } \right| ^2/2\delta _{g}^2 }$ are Gaussian weights assigned according to the spatial distance and the intensity difference. The intensity-guided weights contribute to the spatial smoothness of the 3D frame, which reduces 3D noise but preserves some geometric structure information. This internal summation computes a bilaterally smoothed point in frame $S^T$, which is then mapped back to frame $S^t$ using the integrated motion vectors ${\bf{m}}_i^{t, T}$ (e.g. ${\bf{m}}_i^{t,t + 2}  = {\bf{m}}_i^{t,t + 1}  + {\bf{m}}_i^{t + 1,t + 2}$). In the external summation, $N_t(t)$ is a set of neighboring frames $S^T$ of the frame $S^t$. $\nu_i^{t,T}$ is a flag for the validity of the integrated 3D motion vector from frame $S^t$ to $S^T$. $f(t,T) = e^{-(t - T)^2 /2\delta _{f}^2 }$ is a weight assigned according to the temporal distance. $\kappa _i^T$ and $\kappa _i$ are the cardinalities of the normalization factors for inter-frame fusion and intra-frame filtering respectively. Eq. (5) gives a smoothed 3D point ${\bf{\hat p}}_i^t$ in the frame $S^t$. Overall, both the spatial and temporal piecewise smoothness are guided by the 2D intensity information.

A point ${\bf{p}}_i^t$ without spatial or temporal neighbors is filled with an interpolated point by using its spatial neighboring 3D points as
\begin{equation}
\scalebox{0.7}{${\bf{\hat p}}_i^t  = \left\{ {\begin{array}{*{20}c}
   {\begin{array}{*{20}c}
   {\begin{array}{*{20}c}
   {{\rm{4D}}} & {{\rm{fusion}}} & {{\rm{model}}} & {Eq.(5)}\\
\end{array}} & {\begin{array}{*{20}c}
   {if} & {\begin{array}{*{20}c}
   {{\rm{satisfying}}} & {Eq.(4)}  \\
\end{array}}  \\
\end{array}}  \\
\end{array}}  \\
   {\begin{array}{*{20}c}
   {\frac{1}{\kappa '_i}\sum\limits_{{\bf{p}}_n^t \in N_S({\bf{p}}_i^t )} {h({\bf{p}}_i^t ,{\bf{p}}_n^t )\left( {{\bf{p}}_n^t  + \left\langle {\nabla _{{\bf{p}}_i^t {\bf{p}}_n^t } ,{\bf{p}}_i^t  - {\bf{p}}_n^t } \right\rangle } \right)} } & {otherwise}  \\
\end{array}}  \\
\end{array}} \right.$}
\end{equation}
where $N_S({\bf{p}}_i^t)$ is a set of spatial neighbors of ${\bf{p}}_i^t$, $n$ is the index of the neighbor. $h({\bf{p}}_i^t ,{\bf{p}}_n^t ) = e^{-\left\| {{\bf{p}}_i^t  - {\bf{p}}_n^t } \right\| ^2/2\delta _{h}^2 }$ is the Gaussian weight assigned according to the spatial distance, ${\nabla _{{\bf{p}}_i^t {\bf{p}}_n^t } }$ is the gradient of the neighboring point ${\bf{p}}_n^t$, and $\kappa '_i$ is the cardinality of a normalization factor.

As a result, we can obtain a fused 3D sequence with lower spatial noise and temporal fluctuations.

\section{Results and Discussion}
\label{secresults}
This section presents noise and shape correctness tests on synthetic data and real experiments using a high frame rate 3D sensor to verify the effectiveness and robustness of the proposed algorithm.

\subsection{Synthetic Noise Test}
The synthetic measured object is a falling 3D ball with the radius of $140$~mm. The synthetic 3D sequence contains 50 3D frames. The resolutions of the intensity image and depth image are $600\times600$~pixels and $600\times600$~points respectively. The sphere fell with the speed of $2$~pixels/frame. The roughness of the 3D surface in one frame was measured by averaging (over the central area of the sphere) the local roughness $\Pi_i$ of a 3D point ${\bf{p}}_i^t$ relative to its neighboring patch with the size of $n\times n$ points as
\begin{equation}
\Pi _i  = \sum\limits_j^{n \times n} {\frac{({{\bf{p}}_i^t - {\bf{p}}_j^t) \cdot {\bf{n}}_i }}{{\left| {{\bf{n}}_i } \right|}}} 
\end{equation}
where ${\bf{p}}_j^t$ is the neighboring point in the window around the central point ${\bf{p}}_i^t$, ${\bf{n}}_i$ is the normal vector of the fitted plane of the neighboring points. Note that this form of roughness measure does not have value zero when there is no noise, due to the curvature of the surface. We used the roughness to evaluate the performance because there is no ground truth for the real data experiments and we wanted to be able to compare the simulated and real results using the same measure. 

\begin{figure*}
\centering
\scalebox{1.0}{\includegraphics[width= 13cm]{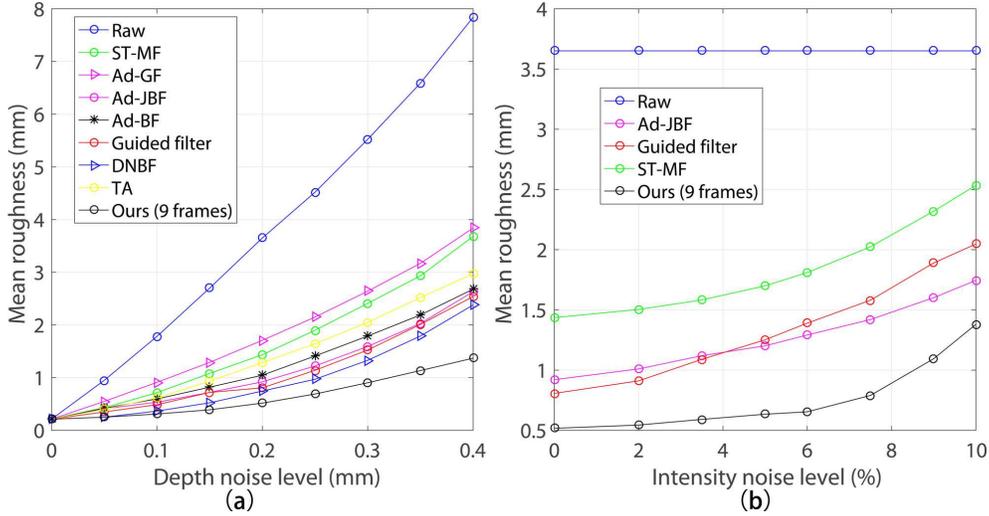}}  
\caption{Mean roughness vs. (a) Depth noise level; (b) Intensity noise level.}
\label{figsynNoise}
\end{figure*}

We added Gaussian random noise with varying noise levels to the intensity and depth images, respectively, and then calculated the mean roughness of the reconstructed 3D sequence. The depth noise level varies from 0.1 mm to 0.4 mm. The intensity values are normalized to [0 1] and the intensity noise level varies from 2\% to 10\% of the highest intensity value. The results were compared with other existing methods including Ad-GF \cite{nguyen2012modeling}, Ad-BF \cite{chen2012depth}, Guided filter \cite{he2013guided}, DNBF \cite{fu2012kinect}, TA \cite{wasza2011real}, Ad-JBF \cite{camplani2013depth}, and ST-MF \cite{matyunin2011temporal}. The mean roughness results (over all frames) w.r.t. different noise levels and algorithms are shown in Fig. \ref{figsynNoise}. 

\begin{figure*}
\centering
\scalebox{1.0}{\includegraphics[width= 13cm]{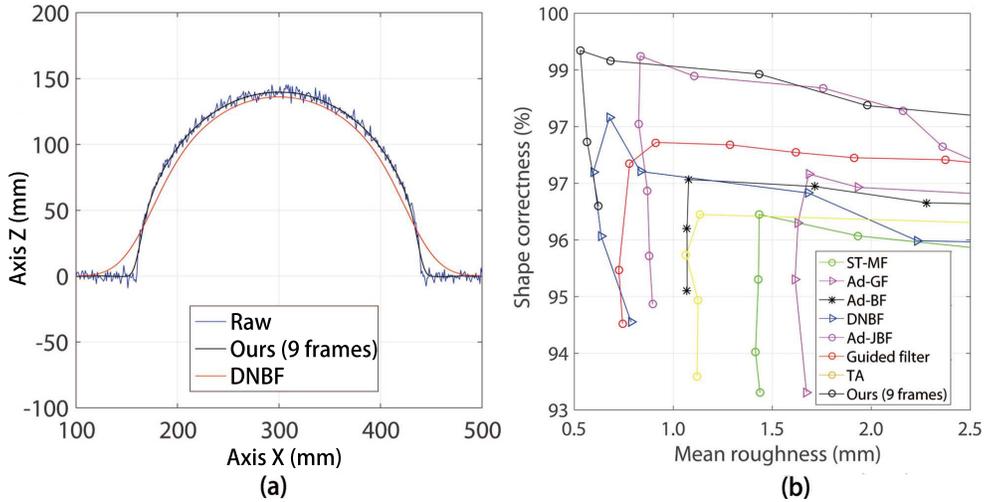}}  
\caption{(a) Illustration of 3D noise reduction on the ball; (b) Roughness vs. shape correctness curves. (The raw data was mean roughness of 3.75 mm and shape correctness of $88.34\%$)}
\label{figbalance}
\end{figure*}

The results in Fig. \ref{figsynNoise}a demonstrate that the performance of the proposed algorithm is superior to other algorithms especially at higher depth noise levels. Some intensity-joint or motion-joint algorithms (Ad-JBF, Guided filtering, DNBF) achieve better results on the synthetic noisy 3D ball than the single image based algorithms such as Ad-BF and Ad-GF. In Fig. \ref{figsynNoise}b, our algorithm has better performance over all the intensity noise levels, followed by the Guided filtering and Ad-JBF. Specifically, for our algorithm, the increments of the mean roughness in lower intensity noise levels are smaller than those in higher levels. This is because the 3D motion vectors are quantized to integral points and some sub-point wrong motion vectors are rejected at the stage of 3D motion field estimation, which increases the robustness of the intensity guided fusion method to some extent. 

\subsection{Roughness vs. Shape Correctness Test}
Roughness and shape correctness are important coupled parameters for describing the quality of 3D reconstructed data. We seek to improve the smoothness of 3D data without losing the shape correctness when oversmoothing happens. In this part, using the falling noisy synthetic sphere (with known ground truth), we investigated the balance between the reduction in roughness and in shape correctness of different algorithms as the amount of smoothing is varied. The results are shown in Fig. \ref{figbalance}. The shape correctness is defined as
\begin{equation}
C = 1 - \frac{{\left| {r - \overline r } \right|}}{{\overline r }}
\end{equation}
where $r$ is the estimated radius of the sphere, computed by the MLESAC algorithm \cite{torr2000mlesac} over data from pixels 160 to 440 (as shown in Fig. \ref{figbalance}a); $\overline r$ is the ground truth radius.

Fig. \ref{figbalance}a illustrates the balance between roughness and shape correctness on the noisy ball from a side view. Our algorithm's smoothed depth values (black curve) have both lower roughness and better shape correctness than the raw values, while the DNBF smoothed depth values (red curve) has worse shape correctness when reaching the same roughness. That means the roughness improvement is achieved by sacrificing some shape correctness, which causes unexpected global deformations of the object. 

For each algorithm, we varied the size of the smoothing neighborhood and the number of smoothing iterations to enable the algorithms to generate different roughnesses and to investigate the corresponding shape correctness. The initial depth noise level is $0.2$~mm and the intensity noise level is 2\%. The quantitative results are shown in Fig. \ref{figbalance}b. Overall, applying different noise reduction algorithms, the mean roughness decreases from the raw roughness (3.75 mm) in different degrees, with increasing shape correctness. However, after the best point, oversmoothing causes serious shape correctness loss with almost the same or even slightly decreasing roughness. Specifically, the curves show that our proposed algorithm achieves the best performance (nearest upper left corner), which demonstrates that it can denoise the 3D data while preserving the structural information better.


\subsection{Results on High Frame Rate Sensors}
\begin{figure}
\centering
\scalebox{1.0}{\includegraphics[width= 8.4cm]{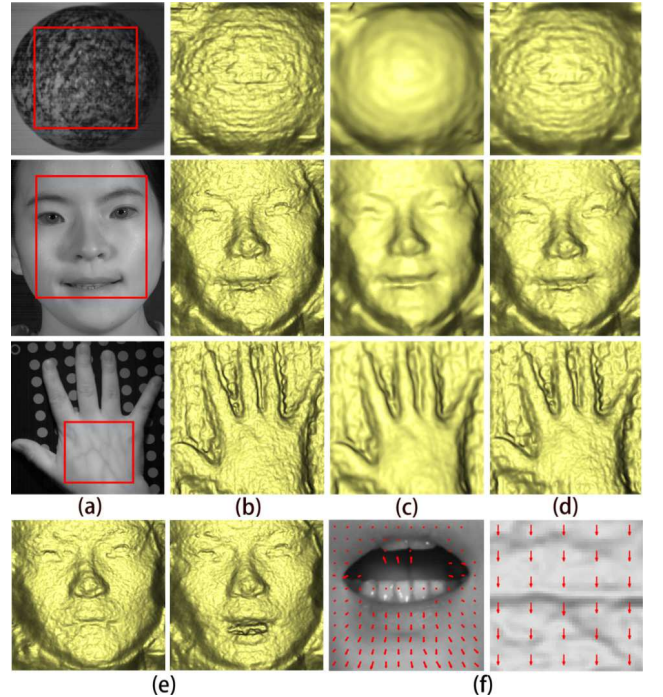}}
\caption{Static Plane (first row): (a) mean roughness; (b) std of roughness vs. number of frames fused. Falling ball (second row): (c) mean roughness; (d) std of roughness vs. number of frames fused} 
\label{figQua}
\end{figure}

\begin{figure*}
\centering
\scalebox{1}{\includegraphics[width= 13cm]{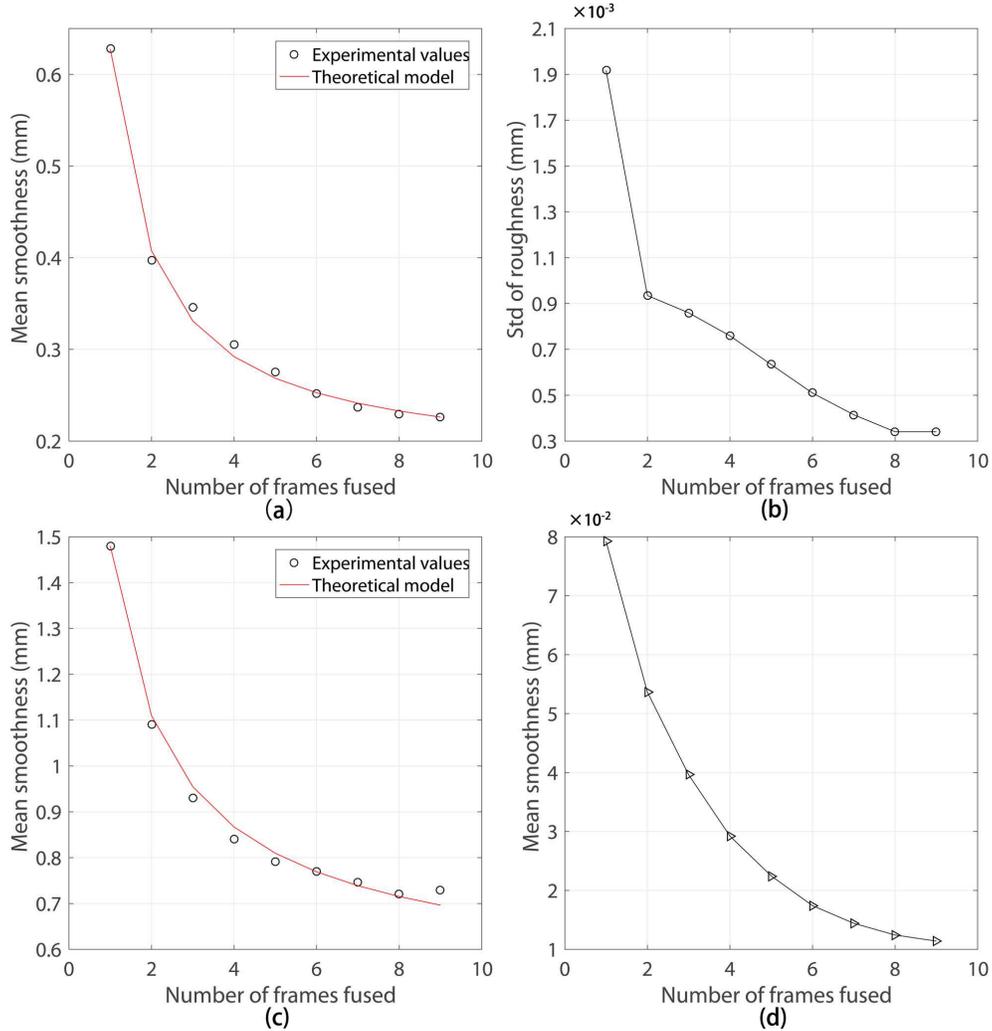}}

\caption{From first to third row: falling ball, dynamic human face, static hand. (a) Intensity frame at time $t$ with ROI marked using a red box; (b) Raw registered 3D frame at time $t$; (c) Improved 3D frame by our algorithm; (d) Improved 3D image by JBF \cite{camplani2013depth}; (e) Raw human face frame at time $t-100$ and $t+100$; (f) Motion field: the mouth region of human face (left) and the center region of falling ball (right).}
\label{fig4}
\end{figure*}

The proposed method was tested on four real 3D objects with different states and surface complexities, including a static plane, a static hand, a falling ball and a speaking human face (as shown in Fig. \ref{figQua}a). The measured stationary plane with textures is $\sim120\times80$~mm. The radius of the ball is $\sim70$~mm. The 3D sequence of the ball is time-varying since the ball deforms and rotates slightly during the falling. For each object, we captured a 3D sequence using a high-speed DI4D system \cite{di4dorg} that consists of a stereo video sensor with the frame rate of $1000$~fps. We applied the proposed method with varying numbers of fused frames to each measured object. For each number of fused frames, we calculated the mean roughness and standard deviation (std) of the 3D sequence. The results are shown in Fig. \ref{fig4}. A qualitative example result of the proposed method when fusing 9 frames is shown in Fig. \ref{figQua}, and the corresponding quantitative comparative results are shown in Table 1.
\begin{table*}
\centering
\caption{Comparative results of 3D/depth noise reduction methods}
\scalebox{1}{
\begin{tabular}{c|cccccccc}
\hline
& \multicolumn{2}{c}{plane $(mm)$} & \multicolumn{2}{c}{hand $(mm)$} & \multicolumn{2}{c}{falling ball $(mm)$} & \multicolumn{2}{c}{dynamic face $(mm)$}\\
& mean & std $(\times 10^{-3})$ & mean & std $(\times 10^{-2})$ & mean & std $(\times 10^{-2})$ & mean & std $(\times 10^{-2})$ \\
\hline
Raw & 0.62	& 1.91&1.34&1.81&1.48&7.92&1.10&6.76\\
Ad-GF \cite{nguyen2012modeling} & 0.39&1.12&0.89&0.41&1.11&3.95&0.77&6.08\\
Ad-BF \cite{chen2012depth} & 0.26&0.82&0.64&1.21&0.89&3.84&0.59&5.45\\
Guided filter \cite{he2013guided} & 0.34&0.73&0.61&1.21&0.83&2.43&0.60&5.09\\
DNBF \cite{fu2012kinect} & 0.32&0.71&0.60&1.18&0.84&2.65&0.58&5.24\\
TA \cite{wasza2011real} & 0.34&0.65&0.71&1.09&0.93&2.41&0.67&4.65\\
Ad-JBF \cite{camplani2013depth} & 0.27&0.61&0.64&1.21&0.89&3.81&0.59&5.45\\
ST-MF \cite{matyunin2011temporal} & 0.36&1.01&0.83&1.38&1.05&3.62&0.73&5.95\\
6D motion field \cite{izadi2011kinectfusion} & 0.39&0.59&0.78&0.91&-&-&-&-\\
6D motion field \cite{dou20153d} &-&-&-&-&0.81&1.97&0.52&\textbf{2.67}\\
\textbf{Ours (9 frames)} & \textbf{0.22}& \textbf{0.31} &\textbf{0.55}&\textbf{0.83}&\textbf{0.71}&\textbf{1.14}&\textbf{0.40}& 2.73\\
\hline
\end{tabular} }
\end{table*}

One can model the mean roughness presented in Fig. \ref{fig4} as $ \sqrt {\delta _s^2  + (1/n)\delta _t^2 } $, where $\delta _s$ is the std of the structural noise, $\delta _t$ is the std of the time-varying noise, and $n$ is the number of frames fused. The red line in Fig. \ref{fig4} and Fig. \ref{fig4} show the above theoretical results fit the experimental results closely. It is obvious that both the mean and std of roughness decrease with the increasing number of frames fused. Compared with the static object, the std of roughness of the dynamic object falls more sharply, when the number of fused images varies from 2 to 9. This is because the number of fused frames mainly influences the temporal dynamic noise, while the dominant noise of the static object is regular structural noise. Overall, we can conclude that the proposed intensity-guided 4D fusion algorithm is more effective and suitable for boosting the 3D reconstruction of dynamic objects.

From the qualitative results in Fig. \ref{figQua} we see that the 3D noise is obviously reduced by ours so that the surfaces of ROIs of the observed 3D objects are much smoother than those in the raw 3D images, especially for the falling ball. Correspondingly, the comparative results in Table 1 demonstrate that our method achieves the best performance with the lowest mean roughness (spatial noise) and the most stable roughness measure (std: temporal fluctuations).

\subsection{In Comparison to 6D Motion Field Based Fusion}
\label{SubsecCompare}
\begin{figure*}
\centering
\scalebox{1}{\includegraphics[width= 13cm]{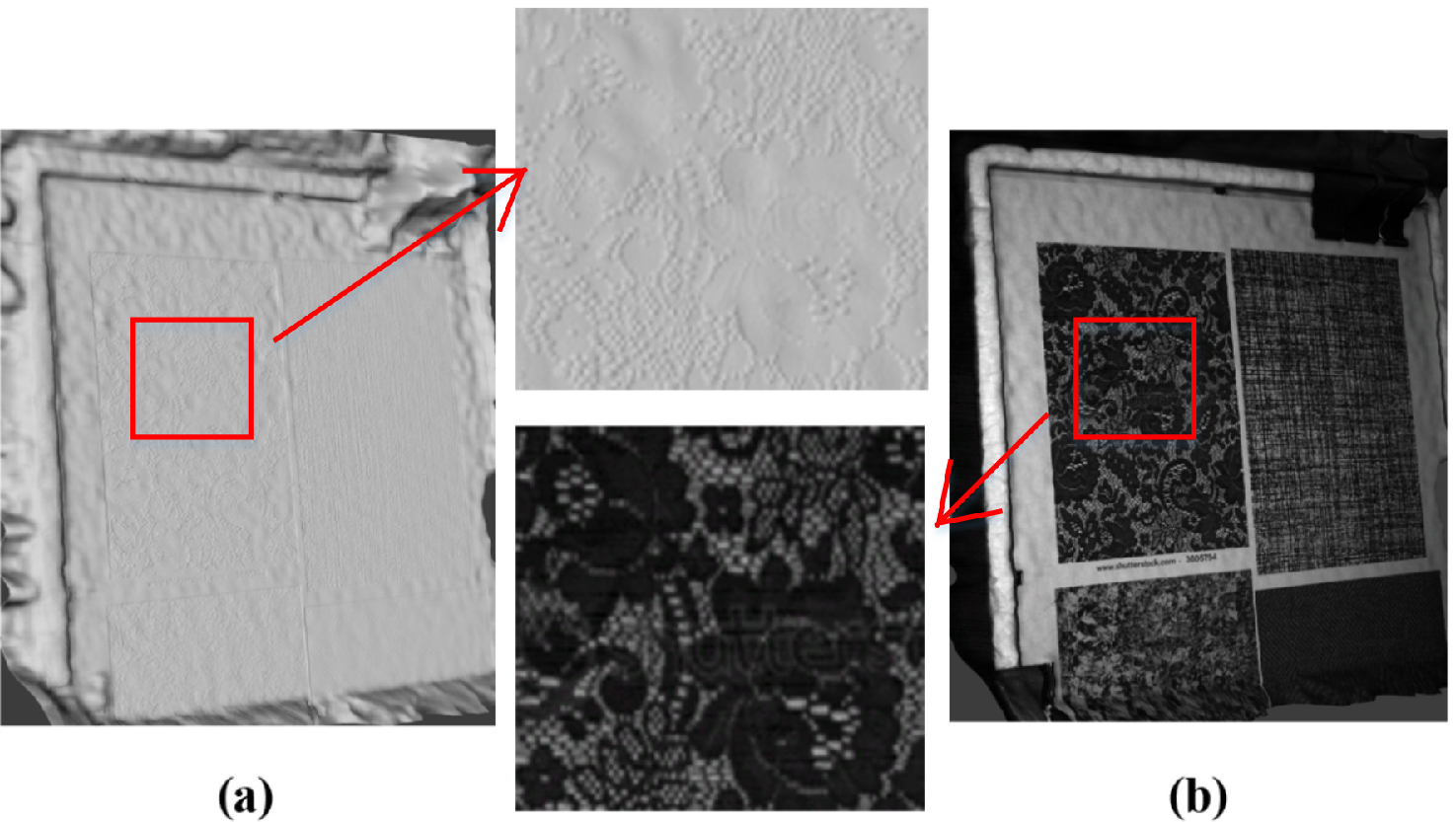}}

\caption{Texture-related 3D noise on a static plane: (a) 3D frame; (b) 3D frame with textures. The 3D noise is closely related to the textures in the intensity image. (better to view online)}
\label{figtexturenoise}
\end{figure*}

In contrast to 4D fusion based on intensity motion fields for 3D/depth noise reduction, there are a group of algorithms that directly generate volumetric 6D motion fields $\{{\bf{R}}_i,{\bf{T}}_i\}$ using depth data from Kinect sensors and reconstruct improved 3D scenes via dense 3D/depth frame registration, such as KinectFusion \cite{izadi2011kinectfusion}, DynamicFusion \cite{newcombe2015dynamicfusion}, 3D Deformable Scanning \cite{dou20153d}, etc. In those works, the multi-view partial 2.5D scans from the Kinect sensors allow for large geometric and pose variations, while our algorithm works on consecutive frames from a 1000 fps 3D video sensor focusing on dense micro-deformation and fusion. Besides, the 3D noise from the 1000 fps video sensor is closely related to the textures of the observed 3D objects due to the uneven reflectance of the textures, as shown in Fig. \ref{figtexturenoise}. Therefore, we directly use intensity information to generate intensity motion fields, guiding the spatio-temporal fusion.

\begin{figure}
\centering
\scalebox{1}{\includegraphics[width= 8.4cm]{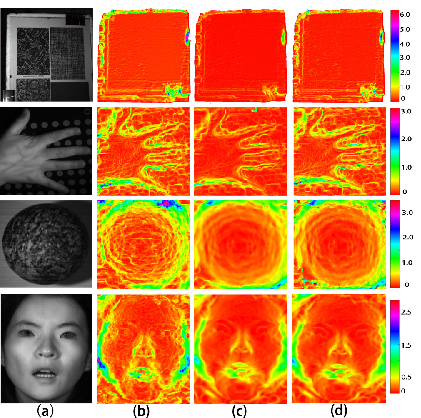}}

\caption{From top to bottom row: static plane, static hand, falling ball, speaking human face. (a) Intensity frame at time $t$; (b) Roughness map of raw registered 3D frame at time $t$; (c) Roughness map of an improved 3D frame by our algorithm; (d) Roughness map of an improved 3D image by 3D motion field based algorithms \cite{izadi2011kinectfusion,dou20153d}. (Better to view online in a color version)}
\label{figfull}
\end{figure}

We compared the performance of the proposed algorithm on the same four objects with the 6D motion field based fusion algorithms. For static objects including the static plane and the hand, a 6D transformation between a pair of consecutive 3D frames was generated using the rigid ICP algorithm, then all the registered 3D points were integrated into a volumetric representation for fusion. For the dynamic and deformable objects including the falling ball and the speaking human face, a dense 6D warp field between pairwise 3D frames was generated using the Embedded Deformable model (ED) based registration method. Then, 9 consecutive frames were fused by leveraging the 8 dense flow fields between each pair of 3D frames. We calculated the roughnesses of the surface of each object and mapped them to the object as shown in Fig. \ref{figfull}. The mean roughness and standard deviation of all 3D frames in a sequence were calculated, as listed in Table 1.

Overall, Both the qualitative results in Fig. \ref{figfull} and comparable results in Table 1 show that our algorithm achieves better results on the datasets. The use of the 2D intensity frames increases the accuracy of dense correspondence and thus improves the spatio-temporal fusion for 3D noise reduction of high frame rate 3D video sensors, especially for the objects with less 3D shape characteristics, such as the plane, hand and ball. Also, our algorithm directly focuses on texture-related 3D noise (Fig. \ref{figtexturenoise}), yielding a texture correspondence guided dense 3D motion field. It is more suitable for high frame rate 3D sequences of dynamic and deformable objects even with fewer 3D shape features.

\section{Conclusion}
This paper presents a simple yet powerful pipeline for improving the 3D reconstruction of dynamic and deformable objects, using 2D intensity tracking guided multi-frame 4D fusion. The continuous motion fields of a 3D sequence are estimated by leveraging the intensity motion fields that are obtained by dense tracking on a pixel-wise registered intensity sequence. Using a spatial-temporal multi-frame 4D fusion model, consecutive 3D frame fusions are performed for improving the spatial smoothness and temporal stability of the 3D sequence. The experimental results on stationary, dynamic and deforming objects verify that the proposed algorithm achieves state-of-the-art performance with the lowest mean roughness over the reconstructed 3D surface in one frame and the best robustness over the whole 3D sequence. In the future, we would like to apply the proposed algorithm as a part of dynamic 3D shape acquisition and recognition (e.g. dynamic 3D human face and hand gesture recognition) to improve the accuracy and robustness of the 3D reconstruction and recognition of highly dynamic and deformable objects.

\ifCLASSOPTIONcompsoc
  \section*{Acknowledgments}
\else
  \section*{Acknowledgment}
\fi

This work is supported by the China Scholarship Council (No. 201606020087), National Council for Science and Technology (CONACyT) of Mexico.

\bibliographystyle{IEEEtran}
\bibliography{bib4Dfusion.bib}

\begin{thebibliography}{10}
\providecommand{\url}[1]{#1}
\csname url@samestyle\endcsname
\providecommand{\newblock}{\relax}
\providecommand{\bibinfo}[2]{#2}
\providecommand{\BIBentrySTDinterwordspacing}{\spaceskip=0pt\relax}
\providecommand{\BIBentryALTinterwordstretchfactor}{4}
\providecommand{\BIBentryALTinterwordspacing}{\spaceskip=\fontdimen2\font plus
\BIBentryALTinterwordstretchfactor\fontdimen3\font minus
  \fontdimen4\font\relax}
\providecommand{\BIBforeignlanguage}[2]{{%
\expandafter\ifx\csname l@#1\endcsname\relax
\typeout{** WARNING: IEEEtran.bst: No hyphenation pattern has been}%
\typeout{** loaded for the language `#1'. Using the pattern for}%
\typeout{** the default language instead.}%
\else
\language=\csname l@#1\endcsname
\fi
#2}}
\providecommand{\BIBdecl}{\relax}
\BIBdecl

\bibitem{xiao2011performance}
Y.~Xiao, R.~B. Fisher, and M.~Oscar, ``Performance characterization of a
  high-speed stereo vision sensor for acquisition of time-varying 3d shapes,''
  \emph{Machine Vision and Applications}, vol.~22, no.~3, pp. 535--549, 2011.

\bibitem{tabata2015high}
S.~Tabata, S.~Noguchi, Y.~Watanabe, and M.~Ishikawa, ``High-speed 3d sensing
  with three-view geometry using a segmented pattern,'' in \emph{Intelligent
  Robots and Systems (IROS), 2015 IEEE/RSJ International Conference on}.\hskip
  1em plus 0.5em minus 0.4em\relax IEEE, 2015, pp. 3900--3907.

\bibitem{zhang2014bp4d}
X.~Zhang, L.~Yin, J.~F. Cohn, S.~Canavan, M.~Reale, A.~Horowitz, P.~Liu, and
  J.~M. Girard, ``Bp4d-spontaneous: a high-resolution spontaneous 3d dynamic
  facial expression database,'' \emph{Image and Vision Computing}, vol.~32,
  no.~10, pp. 692--706, 2014.

\bibitem{SPaction2013}
J.~Wang and Z.~Xu, ``Stv-based video feature processing for action
  recognition,'' \emph{Signal Processing}, vol.~93, no.~8, pp. 2151--2168,
  2013.

\bibitem{SPmotion2015}
J.~Xiang and R.~Liang, ``Motion recognition and synthesis based on 3d sparse
  representation,'' \emph{Signal Processing}, vol. 110, pp. 82--93, 2015.

\bibitem{mallick2014characterizations}
T.~Mallick, P.~P. Das, and A.~K. Majumdar, ``Characterizations of noise in
  kinect depth images: A review,'' \emph{IEEE Sensors journal}, vol.~14, no.~6,
  pp. 1731--1740, 2014.

\bibitem{khoshelham2012accuracy}
K.~Khoshelham and S.~O. Elberink, ``Accuracy and resolution of kinect depth
  data for indoor mapping applications,'' \emph{Sensors}, vol.~12, no.~2, pp.
  1437--1454, 2012.

\bibitem{yu2012shadow}
Y.~Yu, Y.~Song, Y.~Zhang, and S.~Wen, ``A shadow repair approach for kinect
  depth maps,'' in \emph{Asian Conference on Computer Vision}.\hskip 1em plus
  0.5em minus 0.4em\relax Springer, 2012, pp. 615--626.

\bibitem{nguyen2012modeling}
C.~V. Nguyen, S.~Izadi, and D.~Lovell, ``Modeling kinect sensor noise for
  improved 3d reconstruction and tracking,'' in \emph{3D Imaging, Modeling,
  Processing, Visualization and Transmission (3DIMPVT), 2012 Second
  International Conference on}.\hskip 1em plus 0.5em minus 0.4em\relax IEEE,
  2012, pp. 524--530.

\bibitem{park2012spatial}
J.-H. Park, Y.-D. Shin, J.-H. Bae, and M.-H. Baeg, ``Spatial uncertainty model
  for visual features using a kinect™ sensor,'' \emph{Sensors}, vol.~12,
  no.~7, pp. 8640--8662, 2012.

\bibitem{chen2012depth}
L.~Chen, H.~Lin, and S.~Li, ``Depth image enhancement for kinect using region
  growing and bilateral filter,'' in \emph{Pattern Recognition (ICPR), 2012
  21st International Conference on}.\hskip 1em plus 0.5em minus 0.4em\relax
  IEEE, 2012, pp. 3070--3073.

\bibitem{izadi2011kinectfusion}
S.~Izadi, D.~Kim, O.~Hilliges, D.~Molyneaux, R.~Newcombe, P.~Kohli, J.~Shotton,
  S.~Hodges, D.~Freeman, A.~Davison \emph{et~al.}, ``Kinectfusion: real-time 3d
  reconstruction and interaction using a moving depth camera,'' in
  \emph{Proceedings of the 24th annual ACM symposium on User interface software
  and technology}.\hskip 1em plus 0.5em minus 0.4em\relax ACM, 2011, pp.
  559--568.

\bibitem{hasinoff2016burst}
S.~W. Hasinoff, D.~Sharlet, R.~Geiss, A.~Adams, J.~T. Barron, F.~Kainz,
  J.~Chen, and M.~Levoy, ``Burst photography for high dynamic range and
  low-light imaging on mobile cameras,'' \emph{ACM Transactions on Graphics
  (TOG)}, vol.~35, no.~6, p. 192, 2016.

\bibitem{SPregistration}
C.~Zhang, S.~Du, J.~Liu, and J.~Xue, ``Robust 3d point set registration using
  iterative closest point algorithm with bounded rotation angle,'' \emph{Signal
  Processing}, vol. 120, pp. 777--788, 2016.

\bibitem{keller2013real}
M.~Keller, D.~Lefloch, M.~Lambers, S.~Izadi, T.~Weyrich, and A.~Kolb,
  ``Real-time 3d reconstruction in dynamic scenes using point-based fusion,''
  in \emph{3DTV-Conference, 2013 International Conference on}.\hskip 1em plus
  0.5em minus 0.4em\relax IEEE, 2013, pp. 1--8.

\bibitem{essmaeel2012temporal}
K.~Essmaeel, L.~Gallo, E.~Damiani, G.~De~Pietro, and A.~Dipand{\`a}, ``Temporal
  denoising of kinect depth data,'' in \emph{Signal Image Technology and
  Internet Based Systems (SITIS), 2012 Eighth International Conference
  on}.\hskip 1em plus 0.5em minus 0.4em\relax IEEE, 2012, pp. 47--52.

\bibitem{fu2012kinect}
J.~Fu, S.~Wang, Y.~Lu, S.~Li, and W.~Zeng, ``Kinect-like depth denoising,'' in
  \emph{Circuits and Systems (ISCAS), 2012 IEEE International Symposium
  on}.\hskip 1em plus 0.5em minus 0.4em\relax IEEE, 2012, pp. 512--515.

\bibitem{wasza2011real}
J.~Wasza, S.~Bauer, and J.~Hornegger, ``Real-time preprocessing for dense 3-d
  range imaging on the gpu: defect interpolation, bilateral temporal averaging
  and guided filtering,'' in \emph{Computer Vision Workshops (ICCV Workshops),
  2011 IEEE International Conference on}.\hskip 1em plus 0.5em minus
  0.4em\relax IEEE, 2011, pp. 1221--1227.

\bibitem{camplani2013depth}
M.~Camplani, T.~Mantecon, and L.~Salgado, ``Depth-color fusion strategy for 3-d
  scene modeling with kinect,'' \emph{IEEE Transactions on Cybernetics},
  vol.~43, no.~6, pp. 1560--1571, 2013.

\bibitem{he2013guided}
K.~He, J.~Sun, and X.~Tang, ``Guided image filtering,'' \emph{IEEE transactions
  on pattern analysis and machine intelligence}, vol.~35, no.~6, pp.
  1397--1409, 2013.

\bibitem{matyunin2011temporal}
S.~Matyunin, D.~Vatolin, Y.~Berdnikov, and M.~Smirnov, ``Temporal filtering for
  depth maps generated by kinect depth camera,'' in \emph{3DTV Conference: The
  True Vision-Capture, Transmission and Display of 3D Video (3DTV-CON),
  2011}.\hskip 1em plus 0.5em minus 0.4em\relax IEEE, 2011, pp. 1--4.

\bibitem{yang2015evaluating}
L.~Yang, L.~Zhang, H.~Dong, A.~Alelaiwi, and A.~El~Saddik, ``Evaluating and
  improving the depth accuracy of kinect for windows v2,'' \emph{IEEE Sensors
  Journal}, vol.~15, no.~8, pp. 4275--4285, 2015.

\bibitem{besse2014pmbp}
F.~Besse, C.~Rother, A.~Fitzgibbon, and J.~Kautz, ``Pmbp: Patchmatch belief
  propagation for correspondence field estimation,'' \emph{International
  Journal of Computer Vision}, vol. 110, no.~1, pp. 2--13, 2014.

\bibitem{torr2000mlesac}
P.~H. Torr and A.~Zisserman, ``Mlesac: A new robust estimator with application
  to estimating image geometry,'' \emph{Computer Vision and Image
  Understanding}, vol.~78, no.~1, pp. 138--156, 2000.

\bibitem{di4dorg}
``Dimensional imaging (di4d™),'' \url{http://www.di4d.com/}.

\bibitem{dou20153d}
M.~Dou, J.~Taylor, H.~Fuchs, A.~Fitzgibbon, and S.~Izadi, ``3d scanning
  deformable objects with a single rgbd sensor,'' in \emph{Proceedings of the
  IEEE Conference on Computer Vision and Pattern Recognition}, 2015, pp.
  493--501.

\bibitem{newcombe2015dynamicfusion}
R.~A. Newcombe, D.~Fox, and S.~M. Seitz, ``Dynamicfusion: Reconstruction and
  tracking of non-rigid scenes in real-time,'' in \emph{Proceedings of the IEEE
  conference on computer vision and pattern recognition}, 2015, pp. 343--352.

\end{thebibliography}

\ifCLASSOPTIONcaptionsoff
  \newpage
\fi

\end{document}